\documentclass[10pt,twocolumn,letterpaper]{article}

\usepackage{cvpr}              % To produce the CAMERA-READY version
\usepackage[table]{xcolor}

\usepackage{graphicx}
\usepackage{amsmath}
\usepackage{amssymb}
\usepackage{booktabs}
\usepackage{multirow}
\usepackage{pifont}
\usepackage{comment}

\usepackage[pagebackref,breaklinks,colorlinks]{hyperref}

\definecolor{citecolor}{RGB}{34,139,34}

\usepackage[capitalize]{cleveref}
\crefname{section}{Sec.}{Secs.}
\Crefname{section}{Section}{Sections}
\Crefname{table}{Table}{Tables}
\crefname{table}{Tab.}{Tabs.}

\definecolor{darkgreen}{rgb}{0.0, 0.5, 0.0}
\newcommand{\PAR}[1]{\vskip4pt \noindent {\bf #1~}}
\newcommand{\PARM}[1]{\vskip4pt \noindent {\it #1~}}

\newtoggle{comments}
\toggletrue{comments}
\iftoggle{comments}{%
\newcommand{\mat}[1]{{\leavevmode\color{blue}[\textbf{Deva:} #1]}}
\newcommand{\alj}[1]{{\leavevmode\color{cyan}[\textbf{Alj:} #1]}}
\newcommand{\gui}[1]{{\leavevmode\color{orange}[\textbf{Neehar:} #1]}}
\newcommand{\jono}[1]{{\leavevmode\color{purple}[\textbf{Jono:} #1]}}
\newcommand{\lau}[1]{\textcolor{magenta}{\textbf{Laura: }{#1}}}
}
{%
\newcommand{\alj}[1]{}
\newcommand{\mat}[1]{}
\newcommand{\jono}[1]{}
\newcommand{\gui}[1]{}
\newcommand{\lau}[1]{}
}

\newcommand*{\metric}{\textit{forecasting mAP}\@\xspace}
\newcommand*{\apf}{\textit{$AP_{f}$}\@\xspace}
\newcommand*{\ap}{\textit{$AP$}\@\xspace}

\newcommand*{\futuredet}{\textit{FutureDet}\@\xspace}

\def\cvprPaperID{7091} % *** Enter the CVPR Paper ID here
\def\confName{CVPR}
\def\confYear{2022}

\begin{document}

\title{Forecasting from LiDAR via Future Object Detection}

% \title{Multi-Future Forecasting from LiDAR via Future Object Detection}

%\title{Do we need to track to forecast?}

%\title{Multi-modal forecasting from raw sensor data}

% \title{Forecasting via Multi-Modal Future Object Detection}

% \title{Multi-Modal Forecasting via Future Object Detection (directly on LiDAR?)}

% \title{Multi-Future Forecasting from LiDAR via Future Object Detection}

% \title{Forecasting from LiDAR via Future Object Detection}

% \title{Multi-Modal Forecasting Direct from LiDAR via Future Object Detection.}

%Primary focus: do we need 3stage model, or can we do end-to-end?

% Storyline:
% Agree with FaF, this should be done directly from LiDAR - this is important
% Methods that use tracks can do multi-modal (ref papers)
% Existing LiDAR based cant

% We can do multi-modal directly from raw LiDAR data

% Alternartive: Object tracking emerges from forecasting (probably not tho, would put too much emphasis on tracking)

\author{
Neehar Peri\textsuperscript{1\thanks{Work done during an internship at Argo AI}} \enskip Jonathon Luiten\textsuperscript{1,2} \enskip Mengtian Li\textsuperscript{1} \enskip Aljoša Ošep\textsuperscript{1,3} \enskip Laura Leal-Taixé\textsuperscript{3,4} \enskip Deva Ramanan\textsuperscript{1,4} \\
{\small \textsuperscript{1}Carnegie Mellon University  \enskip \textsuperscript{2}RWTH Aachen University \enskip \textsuperscript{3}TUM Munich \enskip \textsuperscript{4}Argo AI} \\
{\tt\small \{nperi,jluiten,mengtial,aosep,deva\}@andrew.cmu.edu, leal.taixe@tum.de}
}
%\thanks{Work done during an internship at Argo AI}

\maketitle

%%%%%%%%% ABSTRACT
\begin{abstract}
% Forecasting the future locations of objects is one of the key computer vision challenges for autonomous vehicles. If systems can predict where objects will be in the near future, they can use this information for safe path planning and collision avoidance.  
% %
% Due to the importance of this task, recent community efforts have been moving towards benchmarking trajectory forecasting models, posed as the problem of predicting future trajectories conditioned on past track observations.  
% %
% However, this approach makes an unrealistic assumption that object detection and tracking are already solved upstream in the pipeline, and forecasting methods can simply be built on top of these results without requiring access to the raw sensor data. 
% %
% This paper argues against this view and poses the forecasting problem as object detection in current and future frames, directly from raw sensor data such as LiDAR streams. This significantly simplifies the model design and leads to higher forecasting accuracy compared to baselines that rely on object trackers and hugh definition maps. 

% [Alternative version by Martin]

Object detection and forecasting are fundamental components of embodied perception. These two problems, however, are largely studied in isolation by the community. In this paper, we propose an end-to-end approach for detection and motion forecasting based on raw sensor measurement as opposed to ground truth tracks. Instead of predicting the current frame locations and forecasting forward in time, we directly predict future object locations and backcast to determine where each trajectory began. Our approach not only improves overall accuracy compared to other modular or end-to-end baselines, it also prompts us to rethink the role of explicit tracking for embodied perception. Additionally, by linking future and current locations in a many-to-one manner, our approach is able to reason about multiple futures, a capability that was previously considered difficult for end-to-end approaches. We conduct extensive experiments on the popular nuScenes dataset and demonstrate the empirical effectiveness of our approach. In addition, we investigate the appropriateness of reusing standard forecasting metrics for an end-to-end setup, and find a number of limitations which allow us to build simple baselines to game these metrics. We address this issue with a novel set of joint forecasting and detection metrics that extend the commonly used AP metrics from the detection community to measuring forecasting accuracy. Our code is available on \href{https://github.com/neeharperi/FutureDet}{GitHub}.
%\alj{I liked the previous version (by Martin) better! Modifications made it longer (and worse) imo. I would suggest to revert.} 
\end{abstract}

%%%%%%%%% BODY TEXT
\section{Introduction}

\begin{figure}[t]
% DEVA: Use \centering instead of \begin{center} to save whitespace
\centering
    \includegraphics[width=0.9\linewidth, trim= 0cm 0cm 7.5cm 0cm, clip]{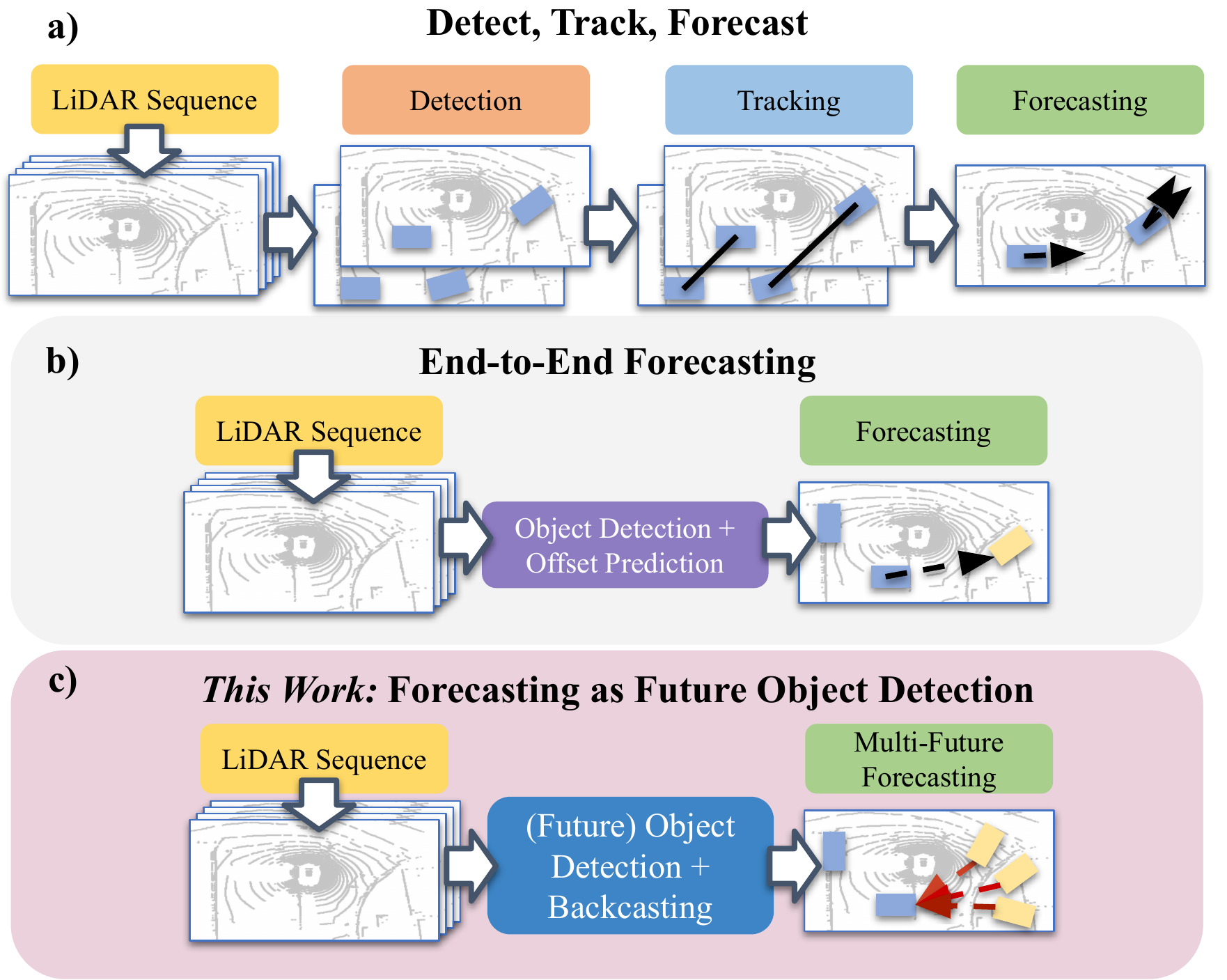}
\caption{\textit{(a)} Current stage-wise methods independently address the problems of detection, tracking, and forecasting, allowing for compounding errors in the full pipeline. Each sub-module incorrectly assumes that its input will be perfect, leading to further integration errors. In contrast to current forecasting methods that use object tracks as input, end-to-end forecasting \textit{directly} from LiDAR sensory data \textit{(b)} streamlines forecasting pipelines. To this end, we propose \futuredet \textit{(c)}, an end-to-end model capable of forecasting multiple-future trajectories directly from LiDAR via future object detection. We show that our end-to-end pipeline improves upon state-of-the-art three-stage and end-to-end methods.}
\label{fig:teaser}
\end{figure}
Object detection and forecasting are fundamental components of embodied perception that are often studied independently.
% This paper is about forecasting. Forecasting is important 
In this paper we rethink the methods and metrics for trajectory forecasting from LiDAR sensor data.
%in terms of both how to approach this task, and how to evaluate it. 
Trajectory forecasting is a critical perception task for autonomous robot navigation, thus building meaningful evaluation metrics and robust methods is of utmost importance.
%We tackle the problem of predicting future locations of objects, one of the most fundamental perception tasks for autonomous robot navigation. Safe future trajectory planning requires understanding how world dynamics evolve over time. 

Traditional trajectory forecasting methods~\cite{Caesar20CVPR, sun20CVPR, chang2019argoverse} detect~\cite{Shi19CVPR, shi20cvpr, shi20pami} and track~\cite{Weng20iros, weng20CVPR, kim21icra} objects in 3D LiDAR scans to obtain past trajectories (Fig.~\ref{fig:teaser}\textit{a}). These can be used in conjunction with auto-regressive forecasting methods~\cite{salzmann2020eccv, ivanovic2018trajectron,  Gupta18CVPR, Alahi16CVPR} to estimate the future actions of surrounding agents.
Recent efforts~\cite{Luo18CVPR, liang2020pnpnet, Weng2021PTP} streamline such multi-stage perception stacks and train multi-task neural networks to jointly detect, track and forecast object positions directly from raw sensor data (Fig.~\ref{fig:teaser}\textit{b}). 
%
% DEVA: I think we can still make a stronger statement about not needing to track, which [31] doesn't claim.
%In particular, Luo~\etal~\cite{Luo18CVPR} jointly detects, tracks, and \textit{forecasts} with a single network operating on an accumulated LiDAR point cloud.  While this is a step in the right direction, this approach predicts
However, such end-to-end approaches tend to predict only a single future trajectory for each object, not accounting for future uncertainty. 
This is not surprising as estimating multiple futures is a significant challenge in forecasting, requiring machinery such as multiple-choice-loss~\cite{bhattacharyya18cvpr} or generative models~\cite{Lee17CVPR, ivanovic2018trajectron,  sadeghian2018sophie,kosaraju2019socialbigat,amirian2019social, Gupta18CVPR, dendorfer2021iccv, Dendorfer2020ACCV}. % or explicitly predicting a distribution over possible future locations~\cite{Dendorfer2020ACCV}. %DEVA: We cite [10] twice. It seems like it fits nicely as a generative GAN model?
%
%Moreover, this method simply regresses relative offsets and, empirically, often fails to predict future positions correctly.
%

We rethink the forecasting task and propose \futuredet, an approach that reframes forecasting as the task of \textit{future object detection} (Fig.~\ref{fig:teaser}\textit{c}). 
%\textit{FutureDet} encodes an accumulated sequence of past raw LiDAR scans using standard backbones for 3D LiDAR-based object detection. 
%
%One of our key insights is that
Importantly, existing detectors~\cite{yin2021center,Lang19CVPR,zhou2018voxelnet} already learn to predict heatmaps that capture distributions over possible object locations. We re-purpose this machinery to represent possible {\em future} object states. 
%
%Based on such a spatio-temporal representation, 
To this end, we encode an accumulated sequence of past raw LiDAR scans using standard backbones for 3D LiDAR-based object detection and  train our network to (i) detect objects multiple timesteps into the future and (ii) estimate trajectories for these future detections back in time (\ie, \textit{back-cast}) to the current frame. % and use these to match future detections with current frame detections. 
By matching back-casted future detections to current detections in a many-to-one manner, our approach can represent a distribution over multiple plausible future states. %, a capability that was previously considered difficult for end-to-end approaches. 
Our extensive evaluation on the large-scale nuScenes~\cite{Caesar20CVPR} dataset for trajectory forecasting reveals that our proposed \textit{FutureDet} outperforms state-of-the-art methods, \textit{without requiring} object tracks or HD-maps as inputs to the model. We posit that tracking may {\em emerge} from our network (since tracking objects from accumulated past LiDAR scans may make them easier to forecast), similar to the emergence of tracking and forecasting in streaming perception~\cite{streaming_li20eccv}.
%
%\deva{This story is shaping up! I'd suggest pointing out that even {\em representing} multiple futures, and the uncertainties between them,  is a notorious challenge in forecasting / future prediction (requiring exotic machinery such as multiple-choice-loss \cite{}, latent-variable sampling~\cite{}, etc.). One of our key insights is to repurpose the machinery of object detection heatmaps to represent multiple modes of possible {\em future} object states (pointing to a figure, perhaps the new teaser?). Finally, I'd also introduce terminonology here; "precast" or "backcast", both of which sound catchy and evocative. Would be nice to put this into the readers head early.} \alj{Addressed this.}

% \begin{itemize}
%     \item Dealing with multi-future nature of the problem: multiple-choice-loss~\cite{bhattacharyya18cvpr}, generative models~\cite{Lee17CVPR, ivanovic2018trajectron,  sadeghian2018sophie,kosaraju2019socialbigat,amirian2019social, Gupta18CVPR, dendorfer2021iccv, Dendorfer2020ACCV}, predicting possible future locations~\cite{Dendorfer2020ACCV}.
%     \item One of our key insights is that we can re-purpose recently introduced CenterPoint~\cite{dendorfer2021iccv} that learns to predict heatmaps that represent a distribution over over possible {\em future} object states.
%     \item Introduce "precast", "backcast"
% \end{itemize}

Furthermore, we investigate the utility of %how useful currently used 
current metrics \cite{liang2020pnpnet, Weng2020SPF2} for evaluating forecasting directly from raw LiDAR data. We find that existing metrics are not well suited for the task of joint detection and forecasting, allowing them to be gamed by trivial forecasters. Current metrics for end-to-end LiDAR forecasting adapt trajectory-based forecasting metrics, such as average/final displacement error (ADE/FDE). These metrics were designed for evaluating forecasting in a setting where perfect tracks are given as input, and objects don't have to be detected. However, these metrics don't adapt well to the end-to-end setting. We demonstrate that such metrics can be gamed by baselines that simply rank all stationary objects (which are trivially easy to forecast) with high confidence, dramatically outperforming all prior art.
% DEVA: Thought we needed more of a gentle intro here
%These metrics were designed for evaluating forecasting in a setting where the perfect trajectories are given as input, and objects don't have to be detected, however they don't adapt well to the end-to-end setting. These adaptions evaluate ADE and FDE \wrt specific recall values \cite{liang2020pnpnet} or average over the recall curve \cite{Weng2020SPF2}, which results in either only a subset of all detections and their forecasts being evaluated, or a subset being weighted much more heavily than others. However these subsets can be selected by the forecaster as they depend on the confidence score, allowing one to easily build a forecaster that can game these metrics into getting a good score by simply scoring easy to forecast objects highly (e.g. those that are static). 
Moreover, these evaluation metrics detach two inherently inter-connected tasks of detection and forecasting. Consequentially, they do not penalize \textit{false forecasts}, \ie, forecasts that do not actually belong to any objects. In this sense, the end-to-end setup and evaluation is more realistic.

%\deva{One more thing to callout is that current forecasting benchmarks fail to evaluate missed forecasts or false positive forecasts arising from missed tracks or false positive tracks. In this sense, our setup is more realistic.}

To address these short-comings, we rethink the evaluation procedure for joint object detection and forecasting directly from sensor data. Our key insight is that the versatile average precision (AP) metric, a gold standard for assessing object detection performance, can be generalized to the task of joint detection and forecasting. The key feature of our novel \metric is that a forecast is correct \textit{only} if the object is both correctly \textit{detected} and \textit{forecasted}. Our \metric is then calculated by simply using the machinery of AP, but using this joint detection and forecasting definition of a true positive.  Furthermore, our \metric can be extended to evaluating multiple-future forecasts for each object by simply evaluating \wrt the top-$K$ most confident forecasts per-detection. 
Our metric appropriately adapts forecasting metrics for end-to-end evaluation: \metric jointly assesses forecasting and detection, penalizing both \textit{missed forecasts} as well as \textit{false forecasts}. It assesses forecasting performance on the full set of object detections and embraces the inherent multi-future nature of forecasting. 
%

% \deva{I suggest we refer to the teaser fig here, to bring it home for the reader.}    

% Longer-term perspectives: It also opens up new research perspectives for others.
\textbf{Contributions:} We (i) repurpose object detectors for the task of end-to-end trajectory forecasting and propose a model that can predict multiple forecasts for each current detection, (ii) rethink trajectory forecasting evaluation and show that detection and forecasting can be jointly evaluated using a generalization of well-accepted object detection metrics, and (iii) thoroughly analyze the performance of our model on the challenging nuScenes dataset \cite{caesar2020nuscenes}, showing that it outperforms both previous end-to-end trainable methods and more traditional multi-stage approaches. 

\section{Related Work}

\PAR{Object Detection and Tracking.} 
Due to recent advances in supervised deep learning~\cite{Krizhevsky12NIPS} and community efforts in dataset collection and benchmarking~\cite{Geiger12CVPR, dendorfer20ijcv, sun20CVPR, Caesar20CVPR, behley2021ijrr}, the research community has witnessed rapid improvement in LiDAR-based 3D object detection~\cite{Lang19CVPR,Shi19CVPR, shi20cvpr, Qi17CVPR_frustum}, tracking~\cite{Weng20iros,yin2021center}, and segmentation~\cite{behley2021ijrr,aygun21cvpr,zhang20CVPR}.
Several methods~\cite{Shi19CVPR, shi20cvpr} follow a well-established two-stage object detection pipeline using point-cloud encoder backbones and a 3D variant of a region proposal network~\cite{Ren15NIPS}, or detect objects as points, followed by classification and bounding box regression~\cite{yin2021center}.
Due to the sparsity of LiDAR point clouds, recent methods accumulate multiple scans over time to improve object detection~\cite{Lang19CVPR, yin2021center} and LiDAR panoptic segmentation performance~\cite{aygun21cvpr}. 
To understand how trajectories of detected objects evolve over time, multi-object tracking methods associate detections using Kalman filters~\cite{Weng20iros}, learned object descriptors~\cite{weng20CVPR,Frossard18ICRA} or regress frame-to-frame offsets~\cite{zhou20ECCV, yin2021center}, typically followed by greedy or combinatorial optimization to resolve ambiguous data association. 
%Method by~\cite{yin2021center} accumulate multiple LiDAR scans to estimate a 1-frame velocity forecast that is used as a cue for data association.
The latter approach can be interpreted as a single frame forecast for track association. However, autonomous vehicles must account for likely future positions of surrounding agents at longer temporal horizons to safely navigate and avoid collisions in dynamic environments.

%However, to safely plan future motion and navigate through dynamic environments,  
%\deva{Depending on our narrative, we could point out that standard multiframe detectors already report velocities, which could be seen as a 1-frame forecast. When talking about this work to people, I like making this point since this suggests our "radical" path isn't that strange.} \alj{Incorporated that thought.}

\PAR{Trajectory Forecasting.} 
Vision-based trajectory forecasting has been posed as the task of predicting future trajectories of agents, given perfect past trajectories and a top-down image (recorded \eg using drones) as input~\cite{pellegrini10eccv,lerner07cgf,Robicquet16eccv}.
Early physics-based models~\cite{social_force} explicitly model agent-agent and agent-environment interactions and have been successfully used to enhance object trackers~\cite{yamaguchicvpr2011,Leal11ICCVW}. 
Recent methods use auto-regressive data-driven models that leverage recurrent neural networks (RNNs) and encoder-decoder-based architectures to encode the past trajectory and estimate its evolution in future frames~\cite{Alahi16CVPR}. 
To deal with the inherent multi-modal nature of the problem, several methods leverage generative models~\cite{goodfellow_gan, kingma14iclr} to learn a distribution over future trajectories~\cite{Lee17CVPR, ivanovic2018trajectron,  sadeghian2018sophie,kosaraju2019socialbigat,amirian2019social, Gupta18CVPR, dendorfer2021iccv, Dendorfer2020ACCV}.
However, these methods and benchmarks tackle forecasting in idealized scenarios, in which the entire visual scene is directly observed, and perfect input trajectories are provided. Both are unrealistic assumptions in automotive and robotics applications. %\item Trajectron++: Fusion of predictions and other data, \eg, semantic maps or ego-motion estimates~\cite{salzmann2020eccv}. 
Due to the importance of forecasting for automotive path planning, recent large-scale automotive benchmarks~\cite{sun20CVPR, Caesar20CVPR, chang2019argoverse} explicitly focus on the trajectory forecasting task. 
Similar to vision-based methods, these benchmarks pose the forecasting problem as trajectory estimation given past trajectories and a high-definition map of the environment. These benchmarks have facillitated a wide suite of HD-map-based forecasting methods, which either represent the HD-map as rasterized images~\cite{chai2019multipath,gilles2021home}, graphs or vectors~\cite{gao2020vectornet,liang2020learning,gu2021densetnt}.
However, both algorithms and the evaluation protocol make the unrealistic assumption that detection and tracking outputs are perfect.
%robots are always localizable within an up-to-date map, and perhaps far more  detection and tracking modules work perfectly. 
%Several methods: detect, track, then forecast (citations needed). However, those multi-stage systems are complex.
% DEVA: I don't think we should (or need to) attack the map as a prior

\PAR{Forecasting from Sensor Data.} Prior methods~\cite{Luo18CVPR, liang2020pnpnet, Weng2021PTP} tackle object detection, tracking, and forecasting jointly by training a single convolutional neural network in a multi-task fashion from accumulated stacks of LiDAR sweeps.   
Alternatively, \cite{Weng2020SPF2} directly forecasts future LiDAR scans and leverages off-the-shelf LiDAR object detectors to detect objects in these forecasted scans. We believe that this approach of end-to-end joint detection and forecasting is a step in the right direction. However, none of the aforementioned end-to-end methods are able to reason about multiple future trajectories. To embrace the inherently uncertain future, we present an end-to-end forecasting model (Sec.~\ref{sec:methodology}) that simultaneously detects objects in the current and future timesteps given a history of LiDAR scans, anchoring multiple possible future detections to current scan detections. This approach not only outperforms the aforementioned methods \wrt forecasting accuracy but also allows for multiple future interpretations. %
Finally, we observe that ad-hoc adaptations of forecasting metrics~\cite{Luo18CVPR, liang2020pnpnet, Weng2021PTP} do not appropriately characterize certain types of forecasting errors. As a remedy, we propose a generalization of the average precision (AP)~\cite{Everingham10IJCV} metric for joint detection and forecasting in Sec.~\ref{sec:experimental_setup}. Note that our adoption of AP is also inspired by the work of streaming perception~\cite{streaming_li20eccv}, where AP is used to measure the joint performance of 2D object detection, tracking, and short-term forecasting without considering multiple futures.
%DEVA: We don't really stress the "lack of tracking". I'm guessing the "tracking as emergent" story never really panned out?

%futures, a capability that was previously considered difficult for end-to-end approaches.
%However, as forecasting raw sensory data is a very difficult problem, this approach is not on-par with state-of-the-art forecasting methods. 

\section{Rethinking Forecasting Evaluation}
\label{sec:experimental_setup}

Since we are tackling the task of forecasting future positions of cars directly from LiDAR scans, we assume an observed sequence of past LiDAR sensor data, up to the most recent observation $\mathcal{S}_{t_{obs}}$ at time $t_{obs}$, as input. 
%\neehar{Good to mention that we accumulate 10 lidar sweeps from the past} \alj{I dont think its relevat at this point, thats goes to method/implementation section.} 
We pose joint object detection and forecasting as the task of estimating a set of object locations (parametrized as 3D cuboids) in the current scan $\mathcal{S}_{t_{obs}}$ as well as their future trajectory continuations in the \textit{future}, unobserved scans \ie, $\{\mathcal{S}_{t}, t \in [t_{obs}+1, \ldots ,T]\}$.

%While there are several possible realizations of future trajectories based on the observed evidence, we can only evaluate performance based on the actually observed trajectories.

%\subsection{Trajectory forecasting}

% In this paper we tackle the task of forecasting future positions of cars and pedestrians directly from the raw LiDAR sensory data. In particular, we assume we observe a sequence of raw LiDAR sensory data, up to the \textit{current} observation at time $t$. 
% We pose joint object detection and forecasting as the task of estimating a set of object detections (parametrized as 2D centroids $x \in \mathcal{R}^2$ on a 2D plane) in the current scan $t$ as well as their future trajectory continuations in \textit{future} scans \ie, $\in [t+1, T]$.
% While there are several possible realizations of future trajectories based on the observed evidence, we can only evaluate performance based on the actually observed trajectories.

\subsection{Preliminaries} 
\label{sec:metric_preliminaries}
Prior work~\cite{pellegrini10eccv,lerner07cgf,Robicquet16eccv, salzmann2020eccv, ivanovic2018trajectron,  Gupta18CVPR, Alahi16CVPR} presents forecasting as the task of estimating the ``correct'' continuation of a given ground-truth track. In particular, given past trajectory observations $X_i = \{(x_i^t, y_i^t) \in \mathbb{R}^2, t=1, \ldots, t_{obs}  \} $, the task is to estimate future positions $Y_i = \{(x_i^t, y_i^t) \in \mathbb{R}^2, t=t_{obs}+1, \ldots, t_{T} \} $ for all agents present in the scene. 
%In addition to past trajectory observations, models are commonly given a top-down image \neehar{not sure what the top down image refers to, maybe good to cite} of the visual scene. 
This formalization is also adopted by recent automotive forecasting benchmarks~\cite{Caesar20CVPR, sun20CVPR, chang2019argoverse}. However, as the ego-vehicle is moving, methods are given high-definition (HD) maps of the surrounding environment and ego-vehicle positions to account for the geometry of the surroundings. First, we discuss existing metrics for forecasting evaluation. 

\PAR{ADE and FDE.} Average displacement error (ADE) and final displacement error (FDE) are commonly used evaluation metrics for assessing trajectory prediction. Both are measured as the L2 distance between model predictions $\{Y_i\}$ and ground truth trajectories $\{G_i\}$. 
To account for the inherent uncertainty in trajectory continuation, methods are evaluated over the set of top-K model predictions (\wrt the confidence score of each predicted forecast). These metrics assume that the set of true positives are the same for all methods. This assumption does not hold when comparing end-to-end methods, which can produce different sets of true positives, making comparison unreliable. 

%\neehar{We should note that these metrics are only comparable between methods if we have a fixed set of true positives. PnP tries to fix this for the end-to-end case where we don't have the same set of detection true positive each time by evaluating at a specific recall threshold. I would put this point here, rather than below.}

\PAR{Miss Rate.} If the final displacement error between a ground-truth trajectory and a prediction is above a center distance threshold, we count the forecast as a miss (similarly evaluated \wrt the set of top-K predictions). This metric evaluates the proportion of misses over all forecasts in a scene. 

\PAR{ADE/FDE \wrt Recall.}
% Adaptations to "real world" (FaF, PnP)
The standard forecasting setup allows us to build models in isolation from other factors and has sparked rapid progress in this field of research~\cite{salzmann2020eccv, ivanovic2018trajectron,  Gupta18CVPR, Alahi16CVPR, Dendorfer2020ACCV, dendorfer2021iccv}. However, the standard assumption of having perfect input trajectories is not feasible in practice as it critically depends on \textit{perfect} object tracks as inputs, which are nearly impossible to obtain in practice.  %, and object tracking is by itself a difficult research problem.  
To this end, \cite{Luo18CVPR, liang2020pnpnet, Weng2021PTP} study end-to-end trajectory forecasting directly from raw sensor data, and propose an evaluation setup for end-to-end forecasting models using the aforementioned ADE and FDE at fixed recall thresholds, \ie, ADE/FDE at 60\% or 90\%. 
This evaluation setting has two major short-comings:\\
\noindent \textit{Evaluated only on a subset of matched detections.} A large number of agents are not moving and prediction of their future positions is trivial. Models that rank stationary objects higher than moving objects can obtain higher forecasting performance by specializing on trivial predictions. We provide empirical evidence for this in Sec.~\ref{sec:breakdown} and show that such recall-based metrics can be ``gamed'' using a simple constant position prediction model. \\
\noindent \textit{No penalty for false positives.} The current evaluation protocol detaches the inter-linked tasks of detection and forecasting. As a consequence, models can predict an arbitrary number of forecasts that are not anchored to any detections. In other words, this approach does not penalize \textit{false forecasts}, \ie, forecasts not anchored to any detection, and \textit{missed forecasts} as commonly characterized by the miss rate.

% \begin{itemize}
%     \item However, these metrics they can be gamed (we can show example of our results, \eg predicting static cars with higher scores).
%     \item They fail to capture properties we actually want from metric (\eg don't encode how good forecasting / detection actually is).    
% \end{itemize}

% \begin{itemize}
%     \item One of the contributions of this paper is re-thinking forecasting evaluation in real-world robotics scenarios. We propose evaluation metrics, suitable for the task of ``end-2-end`` forecasting directly from the sensor data. 
%     %
%     \item This is based on the intuition that forecasting can be posed as the task of future instance detection. For assessing object detection performance in a single (current) scan, we use well-established mean average precision metric (mAP). 
%     %
%     %
%     % %
%     % \item Metrics we propose instead:
%     % \begin{itemize}
%     %     \item f-mAP$_\textrm{f}$ - (forecasting mAp at final) - this is what we said was mAp + missrate. E.g. mAP using the criteria: $\max{(\textrm{current frame error}, \textrm{future frame error})}$.
%     %     \item f-mAP$_\textrm{f}$ - the average of f-mAP$_\textrm{f}$ calculated by setting f to be all frames between 1 and actual f.
%     % \end{itemize}
%     % %
% \end{itemize} 

\subsection{Average Precision is All You Need} 
\label{sec:all-you-need}

%The well-established metric for assessing object detection performance in a single (\ie, current) scan, 
%we use well-established mean average precision metric (mAP). 

% JONO: THE BELOW PARAGRAPH IS A BIG NO NO. IT IMPLIES WE FIT OUR METRIC TO OUR METHOD..... NOOOOO
% In this paper we adopt the view of forecasting as future object detection. Therefore, the natural way to evaluate model performance is to leverage existing approaches for object detection evaluation, including average precision (AP).

% A naive approach would simply evaluate AP in the current scan $\mathcal{S}_{t_{obs}}$ and all future scans $\{\mathcal{S}_{t},  t=t_{obs}+1, \ldots, T \}$ and average the results. However, this approach would (i) not asses detection and forecasting jointly, \ie, false positive forecasts, not associated to any detection would not be penalized \neehar{I don't beleive this is true. The first point about not assessing detection and forecasting still holds.}; \alj{I think its true. We gotta discuss.} and (ii) would not embrace the multi-future nature of the task. The future is inherently uncertain and in the ground-truth data we are only given a single realization of possible trajectories. In the following we discuss how AP can be generalized for the task of forecasting. % and propose \metric (\apf). 
% \deva{FWIW, I find this naive approach not particularly natural. I might suggest starting with the below.}

% Forecasting AP.

\PAR{Average Precision ($AP$).} $AP$ is defined as the area under the precision-recall curve~\cite{Everingham10IJCV}, commonly averaged over multiple spatial overlap thresholds~\cite{Lin14ECCV}.     
To compute \ap we first determine the set of true positives (TP) and false positives (FP) to evaluate precision and recall. 
In standard object detection, TPs are considered to be successful matches between model predictions and ground-truth, typically determined based on 2D/3D intersection-over-union (IoU)~\cite{Everingham10IJCV} or distance from the object center~\cite{Caesar20CVPR} in the reference image or LiDAR point cloud, respectively. We can extend AP for joint detection and forecasting by (a) evaluating detection accuracy on the current frame or (b) evaluating detection accuracy $T$ seconds into the future. However, (a) completely ignores forecasts and (b) doesn't ensure that future trajectories are correctly associated to the right current detection.

\PAR{Forecasting Average Precision ($AP_f$).} For the task of joint detection and forecasting, all future forecasts need to be anchored to objects, present (and detected) in $\mathcal{S}_{t_{obs}}$. A robust metric must correctly penalize trajectories with correct first frame detections and incorrect forecasts (false forecasts), and trajectories with incorrect first frame detections (missed forecasts). 

To characterize both types of errors, we define a true positive with reference to the current frame $t_{obs}$ if there is a positive match in both the current timestamp ($t_{obs}$) \textit{and} the future (final) timestep $t_{obs} + T$. Otherwise, a forecast is considered to be a false positive. 
A successful match in the current frame is determined based on the distance from the center, averaged over distance thresholds of $\{0.5, 1, 2, 4\}$m \cite{caesar2020nuscenes}. Similarly, a successful match in the final timestep is determined based on the distance from object center, averaged over distance thresholds of $\{1, 2, 4, 8\}$m respectively. In contrast to ADE @ Recall \% and FDE @ Recall \%, this evaluation setting (i) takes all detections (not just true positives) into account, and (ii) penalizes missed forecasts (typically characterized by the miss-rate). 

%\deva{This sounds like $AP+MR$, which I'm guessing we are redefining as $AP_f$. If so, I'd point out the link to MR from the forecasting community (which makes this sound more natural). I guess with the new positioning that the old $AP_f$ doesn't make sense, since it doesn't take multiple futures into account. {\em This} line of reasoning might be a more natural intro paragraph - e.g., o...}

\PAR{Forecasting Mean Average Precision ($mAP_f$).} 
\textit{Forecasting AP} is evaluated on the full set of detections and cannot be ``gamed'' by a simple constant position model. However, we note that the data itself is imbalanced: over 60\% of cars in the nuScenes dataset are parked, and are therefore stationary. 
To this end, we define three sub-classes: \textit{static car}, \textit{linearly moving car} and \textit{non-linearly moving car}. Computing sub-class $AP_f$ can be difficult; we do not require forecasts to output sub-class labels, but assume all ground-truth objects have sub-class labels. We follow the protocol for large-vs-small object subclass evaluation from COCO ~\cite{lin2014microsoft}, described further in the appendix. We then evaluate $mAP_f$ as $\frac{1}{3} (AP_f^{\text{stat.}} + AP_f^{\text{lin.}} + AP_f^{\text{non-lin.}})$ to ensure our metric cannot be ``gamed'' by trivial forecasters, as well as discuss fine-grained analysis on the three cases separately. Similarly, $mAP_{det}$ is evaluated as the average $AP_{det}$ over the three sub-classes. 

\PAR{Metrics: Embracing Multiple Futures.}
As described, $mAP_f$ would be suitable for evaluating forecasting for scenes with multiple future ground truth trajectories. However, this is not feasible in the practice when forecasting directly from historical sensor data. To this end, we adopt a top-K based forecasting evaluation~\cite{pellegrini10eccv,lerner07cgf,Robicquet16eccv}, that does not penalize models for hypothesizing possible future trajectories anchored from a single detection. 
In particular, we first match predictions to ground-truth detections in $t_{obs}$ and take the top-K highest ranked forecasts for each detection. Based on this set, we determine the best-matching forecast in terms of FDE and evaluate \apf as explained above.

    % \item Take top-K predictions based on the confidence
    % \item Iterate over GT trajectories, find which one best matches the predicted trajectory (in terms of FDE, throw away others)
    % \item Result: filtered set of GT and predictions, run "evaluation" 
    % \item Our metric: AP+miss_rate
    % \begin{itemize}
    %     \item Match in the current timestamp, 1 AND if they match Match in the final timestamp
    %     \item Count as 0 if it does not match either in current or final time step (all other premutations are zero)
    %     \item Compute PR curve
    % \end{itemize}

%     % \item A true positive: positive match in the in the current timestamp ($t_{obs}$) and final timestep $T$. Otherwise it is a FP.
%     % \item Computing AP this way will (lalala). 
%     % \item However, this approach assumes only one possible outcome. 
%     % %
%     % \begin{itemize}
%     %     \item 
%     %     \item Count as 0 if it does not match either in current or final time step (all other premutations are zero)
%     %     \item Compute PR curve
%     % \end{itemize}    

% % Item why are our metrics better and how they do not suffer from aforementioned issues. 
% \noindent Why our metrics are better? 
% \begin{itemize}
%     \item They actually enable end-to-end forecasting from raw sensor data, \ie, they measure what's important. 
%     \item This also enables prediction of multiple futures, and naturally deals with it by enabling ranking of predictions.
% \end{itemize}
% %%%%%%%%%%%%%%%%%%%%%%

%%%%%%%%%%%%%%%%%%%%%%%%%%%%%%%%%%%%%%%%
% \section{Rethinking forecasting methodology}
\section{Forecasting as Future Object Detection}
\label{sec:methodology}
\begin{figure*}[ht!]
\begin{center}
    \includegraphics[width=0.9\linewidth, trim= 0cm 7cm 2cm 0cm, clip]{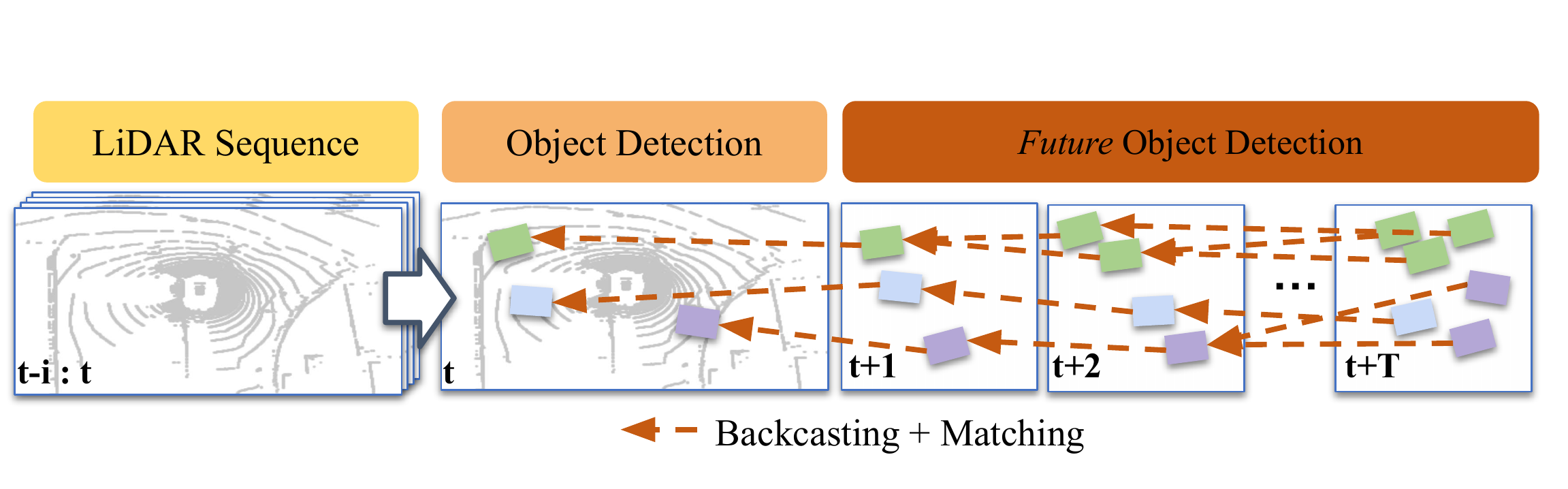}
\end{center}
\vspace{-20pt}
\caption{\textbf{\textit{FutureDet.}} Based on an accumulated LiDAR sequence, \textit{FutureDet} detects objects in the \textit{current} frame $t$ and in a \textit{future} frames up to $t+T$. 
%We forecast via future detection by accumulating and encoding a sequence of LiDAR points to detect objects in the \textit{current} and several \textit{future} time-steps. 
We then cast these future detections back-in-time (\ie, \textit{back-cast}) to the current-frame where they are matched to current frame object detections. Such matching of multiple future detections to current-frame detections is a a natural mechanism for a multi-future interpretation of the observed evidence.}
\label{fig:method}
\end{figure*}

\futuredet addresses the forecasting problem by predicting the future locations of objects observed at $t_{obs}$. We can repurpose existing LiDAR detectors to predict object locations for $T$ future (unobserved) LiDAR scans, for which ground truth supervision is given. We first describe our method and discuss our implementation based on the recently proposed CenterPoint LiDAR detector.~\cite{yin2021center}.
% directly using ground-truth to be the set of detections N frames in the future (where N is maximum number of frames desired to be forecasted), instead of the detections in the current frame. 

%\PAR{From future detections to forecasting.} 

Future object detection and forecasting are related tasks. Forecasting requires predicting consistent trajectories in every frame between the current frame and $T$ future frames. 
To estimate forecasts from future detections we train our network to additionally estimate velocity offset vectors for every future detection. We do so for all frames between the current timestep and the final future timestep where the future detection occurs. 
%We can regress offsets for each future detection, where each offset is trained to estimate the relative offset between frames, \eg, $(N, N-1), (N-1, N-2), \ldots, (1, 0)$. 
%
\PAR{Backcasting vs. Forecasting.} Fast and Furious~\cite{Luo18CVPR} proposes a similar architecture that \textit{forecasts} position offsets into the future directly from current-frame detections. Our method considers the inverse setup where we detect in both the current and future frames and predict offsets back in time. We posit that future object detection requires the network to learn forecasted feature representations ~\cite{luc2018predicting}, directly optimizing for future object positions. Our experimental evaluation and visual inspection confirm this intuition (Sec.~\ref{sec:experimental_setup}). 

\PAR{Method: Embracing Multiple Futures.} The task of forecasting is inherently ambiguous: while there are many plausible outcomes given an input trajectory, only one future is realized for training supervision and evaluation. 
Traditional forecasting methods and benchmarks facilitate multiple future predictions via top-K based evaluation, leveraging multiple-choice-loss~\cite{bhattacharyya18cvpr} and generative models~\cite{Lee17CVPR, ivanovic2018trajectron,  sadeghian2018sophie,kosaraju2019socialbigat,amirian2019social, Gupta18CVPR, dendorfer2021iccv, Dendorfer2020ACCV} to learn a (possibly multi-modal~\cite{dendorfer2021iccv, Dendorfer2020ACCV}) distribution over future trajectories. 
%
%Existing forecasting methods adopt this philosophy for designing models~\cite{} and model evaluation~\cite{}. %Traditionally, forecasting approaches deal with this ambiguity by predicting multiple future forecasts instead of only a single one. 
%
\futuredet allows for natural multi-future forecasting to emerge. We first point out that detection networks can be easily repurposed for future detection by giving target bounding boxes $T$ seconds into the future. 
%This should already exhibit multiple future predictions or ``multiple bets where objects may end up". 
Since future objects are detected independently from current-frame detections, we posit that the network will produce multiple future detections for every object in the scene, effectively placing ``multiple bets'' where the objects may end up in the future. As all future detections are modeled by Gaussian heat maps, we implicitly obtain a multi-modal distribution over possible future locations (see Fig.~\ref{fig:method}).

\PAR{Matching Multiple Forecasts.} The task of forecasting requires all trajectories to be anchored to the set of object detections in the current (observed) LiDAR scan. 
%To achieve this, we first (Fig.~\ref{fig:method}a) use our detection heads to detect objects in current frame. This gives us the number of actual objects in the scene at the current timestep and their locations, modelled as Gaussians. In parallel, we detect (possibly multiple) objects in the future frames (Fig. \ref{fig:method}b). 
%We then match every future detection to a current detection via \textit{backcasting} as follows (Fig. \ref{fig:method}c). We find the closest current detection to the final \textit{backcasted} location for each of the future detections. Finally, we assign a consistent identity to matched current- and future-frame detections. 
For every future detection $i$, we backcast and compute the distance to each detection $j$ from the previous timestep. For each $i$, we pick the best $j$ (allowing for many-to-one matching). 
This framework naturally allows potentially multiple future forecasts to belong to each current timestep detection.

\PAR{Ranking Multiple Forecasts.} For all forecasts anchored at a single detection, we rank trajectories according to their forecasting score, derived using the detection confidence score of the last detection in a predicted trajectory. As shown in Table \ref{table:results}, we find there is a slight increase in performance between $K=1$ and $K=5$, indicating that better ranking strategies can further improve \futuredet.

\PAR{Implementation.} We train CenterPoint to detect objects in future scans. The underlying detection network simply thinks it's finding $T$ times as many object classes (e.g., {\tt cars} and {\tt future-cars}) with additional regression offsets (analogous to existing velocity regressors). In addition, we repurpose the ground truth sampling (\textit{a.k.a.} copy-paste) augmentation \cite{zhu2019class} to increase the diversity of training trajectories. This provides considerable improvement in linear and non-linear forecasting performance.
%\alj{I assume network is given the time=step info too? Also, jono, notation!} %
%Each detector head outputs a number of attributes, including object size, orientation and current velocity. We simply add $N$ additional velocity attributes to be predicted, which are the relative `back-cast' offsets from the future detection back to the current timestep. 
We use the PyTorch toolbox to train all models for 20 epochs with the Adam optimizer and a one-cycle learning rate scheduler.

\PARM{CenterPoint is already a one-frame forecaster.} It detects objects and predicts one-frame future velocity vectors that are used as cues for tracking. It does this by accumulating $T$ previous LiDAR scans and encodes the accumulated point cloud sequences using a VoxelNet~\cite{zhou2018voxelnet} backbone. Such a tracker could be used as input to auto-regressive forecasting methods, \eg, \cite{salzmann2020eccv}, however, we argue that we can use such a spatiotemporal representation to directly forecast. 
\PARM{CenterPoint models object locations as Gaussians.} It does so by producing a 2D bird's-eye-view (BEV) heatmap, which models the continuous likelihood of detections at each point in the BEV space. 
Detections are obtained by finding local maxima in these heatmaps via non-maximum suppression (NMS). 
By reusing this representation for future detection, our detection heatmaps are effectively a forecast of a continuous likelihood field for the locations of objects. This continuous field naturally encodes the uncertainty of future detections, accounts for multi-modality, and provides a continuous representation for forecasting. 
\section{Experimental Evaluation}

%---------------------------------------
% What do we show in exp. sec.?
%

We conduct our experimental analysis on the nuScenes~\cite{Caesar20CVPR} dataset. 
As we are tackling end-to-end forecasting from sensor data, we do not follow the established evaluation protocol that provides ground-truth trajectories and HD maps as input (as explained in  Sec.~\ref{sec:metric_preliminaries}).
First, we perform breakdown analysis of evaluation metrics proposed in~\cite{liang2020pnpnet} and our \metric by analyzing the performance of a simple constant position model (Sec.~\ref{sec:breakdown}). 
After verifying that our proposed evaluation setting is not easily ``gamable'', as discussed in Sec.~\ref{sec:metric_preliminaries}, we thoroughly ablate our model and compare its performance to other state-of-the-art methods (Sec.~\ref{sec:ablation}).

%---------------------------------------
% Talk about the dataset
%
\PAR{Repurposing NuScenes Tracking Dataset.}  
nuScenes~\cite{Caesar20CVPR} recently introduced a large-scale multi-modal dataset recorded in Boston and Singapore. It provides 1000 twenty-second logs that are fully annotated with 3D bounding boxes. This work explicitly focuses on forecasting based on LiDAR data, obtained with a 32 beam LiDAR sensor recorded at 20 Hz, covering 360-degree view.
We follow the official protocol and evaluate forecasting on the \textit{car} class up to 3 seconds in the future. We evaluate forecasting performance on the \textit{pedestrian} class in the appendix. As the test set is hidden, we follow~\cite{liang2020pnpnet} and conduct our analysis on the official validation split.

%---------------------------------------
% Metric (breakdown) analysis
%
\subsection{Metric Breakdown Analysis}
\label{sec:breakdown}

\begin{table*}[htbp]
    \centering
    % \scriptsize
\setlength{\tabcolsep}{0.8pt}
% \renewcommand{\arraystretch}{0.4}
   % \scriptsize
      \resizebox{\linewidth}{!}{%
    %%%%%%%%%%%%%%%%%%%%%%
    %%%%%%%%%%%%%%%%%%%%%%
\begin{tabular}{lccccccccccc}
\toprule 
 & $ADE@60 \;(\downarrow)$ & $FDE@60 \;(\downarrow)$ & $ADE@90 \;(\downarrow)$ & $FDE@90 \;(\downarrow)$ & $ADE  \;avg.\;(\downarrow)$ & $FDE  \;avg.\;(\downarrow)$ &  $AP_{f}^{stat.} (\uparrow)$ & $AP_{f}^{lin.} (\uparrow)$ & $AP_{f}^{non-lin.} (\uparrow)$ & $mAP_{f} (\uparrow)$\tabularnewline
\midrule 
\rowcolor{blue!10} Constant Position (CP) &  \textbf{0.38} & \textbf{0.63} & \textbf{0.48} & \textbf{0.76} & \textbf{0.37} & \textbf{0.64} & \textbf{66.3} & 0 & 0 & 22.1 \tabularnewline
PnPNet~\cite{liang2020pnpnet} &  0.58 & 0.93 & 0.68 & 1.04 & - & - & - & - & - & - \tabularnewline
PnPNet w/o Tracker~\cite{liang2020pnpnet} & 0.69 & 1.09 & 0.75 & 1.14 & - & - & - & - & - & -\tabularnewline
Trajectron++~\cite{salzmann2020eccv} & 1.13 & 2.54 & 1.25 & 2.71 & 1.08 & 2.42 & 59.2 & 8.1 & 2.8 & 23.4\tabularnewline
SPF2~\cite{Weng2020SPF2} &  - & - & - & - & 1.04 & 1.04 & - & - & - & -  \tabularnewline
Fast and Furious$^*$ (FaF)~\cite{Luo18CVPR} & 0.74 & 1.59 & 0.83	& 1.69 & 0.73	& 1.56 & 64.8 & \textbf{22.2} & \textbf{7.5} & \textbf{31.5} \tabularnewline
%Future Det. + Backcast (Ours) & \textbf{0.37} & \textbf{0.56} & 0.83 & 1.56 & \textbf{0.40} & \textbf{0.66} & \textbf{58.6} & \textbf{85.7} & 8.7 & 47.2 \tabularnewline
%Future Det. + Backcast + Match (Ours) & 0.73 & 1.48 & 0.87 & 1.67 & 0.72 & 1.46 & 54.7 & 84.8 & \textbf{16.0} & \textbf{50.4} \tabularnewline
\bottomrule
\end{tabular}

    %%%%%%%%%%%%%%%%%%%%%%
    %%%%%%%%%%%%%%%%%%%%%%
    }
    \caption{\textbf{Metric Breakdown Analysis:} We compare our simple constant position model to state-of-the-art prediction models, highlighting differences among various proposed metrics. ADE/FDE based metrics measured at different recall rates favor our trivial constant position baseline over state-of-the-art methods~\cite{liang2020pnpnet,Luo18CVPR,salzmann2020eccv}. Only our proposed \metric ($mAP_f$) favors state-of-the-art models over the constant position baseline. 
    %We note that Future Det. + Backcast is the only model that beats the CP baseline. Intuitively, backcasting ranks stationary objects higher since they are in the same location in both the current and future frames.
    We report numbers for PnPNet~\cite{liang2020pnpnet} and SPF2~\cite{Weng2020SPF2} from their respective papers. 
    \textbf{Note: Lower ADE/FDE is better and higher $AP_f$ is better.}}
    \label{tab:breakdown}
\end{table*}

In this section, we analyze different evaluation metrics by comparing a trivial constant position model to several state-of-the-art forecasting methods~\cite{liang2020pnpnet,Luo18CVPR,salzmann2020eccv}. 
For the task of end-to-end forecasting from raw data, methods report both detection and forecasting confidence scores. For the simple constant position model, we threshold trajectories such that we only report those where the final position overlaps with the initial position (i.e the object is stationary) with high confidence.
A good forecasting evaluation metric should indicate that the trivial constant position baseline is not a good predictor, as it only correctly predicts future locations of stationary objects and explicitly assumes a stationary world. Do existing metrics reveal this? 

To answer this question, we report the results of our analysis in Table~\ref{tab:breakdown}. 
We analyze the results using average and final displacement (ADE and FDE) errors at $\{60, 90\}\%$ recall~\cite{liang2020pnpnet}, and a variant that averages results over all recall thresholds~\cite{Weng2020SPF2} (see Sec.~\ref{sec:metric_preliminaries} for a discussion of these metrics). 
Our trivial constant position baseline yields state-of-the-art results under the aforementioned evaluation settings. Current metrics are ``gameable'' by this trivial forecaster. 

What about our \metric (Sec.~\ref{sec:all-you-need})? We analyze these methods both through the lens of each motion class ($AP_f^{stat.}$, $AP_f^{lin.}$ and $AP_f^{non-lin.}$), and as an aggregate. 
First, we observe that the constant position model $AP_f$ evaluated over static cars performs better than FaF*, a state-of-the-art end-to-end forecaster. However, when we evaluate on the subset of cars that are in motion, 
%We observe that this is not due to the issue with metric, it is due to the data distribution. 
%To evaluate this hypothesis, we evaluate \metric on subset cars that are in the state of motion. 
$AP_f^{lin.}$ and $AP_f^{non-lin.}$ confirms that our metric behaves as expected: we obtain $0AP$ with the constant position model, indicating that it fails to predict the motion of moving cars. On the other hand, FaF$^{*}$ obtains 7.5 $AP_f^{non-lin.}$, indicating that motion forecasting from the raw sensory data is a very challenging problem. We observe that Trajectron++ outperforms the constant position model for moving objects (8.1  $AP_f^{lin.}$), but does not reach the performance of the constant position model or FaF$^{*}$ on stationary objects.

A good metric should summarize the performance on the full set of cars, \ie, in addition to predicting the motion of moving cars, a good model should correctly predict that parked cars are unlikely to move in the near future. 
This is achieved by our \metric that averages $AP_f^{stat.}$, $AP_f^{lin.}$ and $AP_{non-lin.}$. Our $mAP_f$ ranks state-of-the-art FaF$^{*}$ (31.5 $mAP_f$) favourably over the constant position baseline (22.1 $mAP_f$), as expected. 
%It, however, it favors CP over Trajectron++, which is not surprising: while Trajectron++ performs better on non-linear cars \neehar{XXX} $AP_f^{lin.}$, it under-performs on stationary (\neehar{XXX} $AP_f^{stat.}$), which (intuitively) yields overall lower $mAP_f$ of \neehar{XXX}.
Based on this analysis, we are confident we have the right tools to analyze \textit{FutureDet} thoroughly!

%---------------------------------------
% Model: build-up from simple baselines
%
\subsection{Ablation and Comparison to State-of-the-Art}
\label{sec:ablation}

\begin{table*}[ht!]
\centering
   % \scriptsize
%   \setlength{\tabcolsep}{0.8pt}
   \setlength{\tabcolsep}{01.5pt}
      \resizebox{\linewidth}{!}{%

\begin{tabular}{lccccccccccccccccccccccc}
\toprule 
 & \multicolumn{11}{c}{K=1} &  & \multicolumn{11}{c}{K=5}\tabularnewline
\cmidrule{2-12} \cmidrule{3-12} \cmidrule{4-12} \cmidrule{5-12} \cmidrule{6-12} \cmidrule{7-12} \cmidrule{8-12} \cmidrule{9-12} \cmidrule{10-12} \cmidrule{11-12} \cmidrule{12-12} \cmidrule{14-24} \cmidrule{15-24} \cmidrule{16-24} \cmidrule{17-24} \cmidrule{18-24} \cmidrule{19-24} \cmidrule{20-24} \cmidrule{21-24} \cmidrule{22-24} \cmidrule{23-24} \cmidrule{24-24} 
 & \multicolumn{2}{c}{$AP^{stat.}$} &  & \multicolumn{2}{c}{$AP^{lin.}$} &  & \multicolumn{2}{c}{$AP^{non-lin.}$} &  & \multicolumn{2}{c}{$mAP$} &  & \multicolumn{2}{c}{$AP^{stat.}$} &  & \multicolumn{2}{c}{$AP^{lin..}$} &  & \multicolumn{2}{c}{$AP^{non-lin.}$} &  & \multicolumn{2}{c}{$mAP$}\tabularnewline
\cmidrule{2-3} \cmidrule{3-3} \cmidrule{5-6} \cmidrule{6-6} \cmidrule{8-9} \cmidrule{9-9} \cmidrule{11-12} \cmidrule{12-12} \cmidrule{14-15} \cmidrule{15-15} \cmidrule{17-18} \cmidrule{18-18} \cmidrule{20-21} \cmidrule{21-21} \cmidrule{23-24} \cmidrule{24-24} 
 & $\textrm{AP}_{\textrm{det.}}$ & $\textrm{AP}_{\textrm{f}}$ &  & $\textrm{AP}_{\textrm{det.}}$ & $\textrm{AP}_{\textrm{f}}$ &  & $\textrm{AP}_{\textrm{det.}}$ & $\textrm{AP}_{\textrm{f}}$ &  & $\textrm{mAP}_{\textrm{det.}}$ & $\textrm{mAP}_{\textrm{f}}$ &  & $\textrm{AP}_{\textrm{det.}}$ & $\textrm{AP}_{\textrm{f}}$ &  & $\textrm{AP}_{\textrm{det.}}$ & $\textrm{AP}_{\textrm{f}}$ &  & $\textrm{AP}_{\textrm{det.}}$ & $\textrm{AP}_{\textrm{f}}$ &  & $\textrm{mAP}_{\textrm{det.}}$ & $\textrm{mAP}_{\textrm{f}}$\tabularnewline
\midrule 
% Detection + Constant Position (CP) & 62.0 & 54.0 & &  88.6 & 84.8  & & 3.6 & 0.0  & & 46.1 & 42.2  & & 62.0 & 54.0 & & 88.6 & 84.8 & & 3.6 & 0.0 & & 46.1 & 42.4  \tabularnewline
%\midrule 
Detection + Constant Velocity & \textbf{70.3} & \textbf{66.0} &  & 65.8 & 21.2 &  & 90.0 & 6.5 &  & \textbf{75.4} & 31.2  &  & 70.3 & 66.0 &  & 65.8 & 21.2 &  & 90.0 & 6.5 &  & \textbf{75.4} & 31.2  \tabularnewline
%\midrule 
Detection + Forecast (\cf~\cite{Luo18CVPR}) & 69.1 & 64.7 &  & \textbf{66.1} & 22.2 &  & 86.3 & 7.5 &  & 73.8 & 31.5  &  & 69.1 & 64.7 &  & \textbf{66.1} & 22.2 &  & 86.3 & 7.5 &  & 73.8 & 31.5\tabularnewline
%\midrule
Trajectron++~\cite{salzmann2020eccv}  & \textbf{70.3} & 59.2 &  & 65.8 & 8.1 &  & 90.0 & 2.8 &  & \textbf{75.4} & 23.4  &  & 70.3 & 61.7 &  & 65.8 & 9.8 &  & 90.0 & 4.3  &  & 75.4 & 25.3 \tabularnewline
\midrule 

\rowcolor{green!10} \textbf{\futuredet}  & 70.1 & 65.5 &  & 62.9 & \textbf{24.9} &  & 91.8 & \textbf{10.1} &  & 74.9 & \textbf{33.5}  &  & 70.1 & 67.3 &  & 62.9 & 27.7 &  & 91.7 & 11.7  &  & 74.9 & 35.6  \tabularnewline
\rowcolor{green!10} \futuredet-PointPillars  & 70.1 & 64.1 &  & 63.4 & 24.8 &  & \textbf{92.4} & 9.6 &  & \textbf{75.4} & 32.8  &  & \textbf{70.7} & \textbf{67.5} &  & 63.4 & \textbf{28.8} &  & \textbf{92.0} & \textbf{11.9}  &  & \textbf{75.4} & \textbf{36.1} \tabularnewline
\rowcolor{green!10} \futuredet + Map  & 70.2 & 65.5 &  & 62.7 & 24.3 &  & 91.7 & 9.4 &  & 74.9 & 33.1  &  & 70.2 & \textbf{67.5} &  & 62.7 & 27.1 &  & 91.7 & 11.0  &  & 74.9 & 35.2 \tabularnewline

% \midrule 
\bottomrule
\end{tabular}
}%

%
% Constant position (CP) &  &  &  &  &  &  &  &  &  &  &  &  &  &  &  &  &  &  &  &  &  &  & \tabularnewline
% \midrule 
% Constant velocity (CV) &  &  &  &  &  &  &  &  &  &  &  &  &  &  &  &  &  &  &  &  &  &  & \tabularnewline
% \midrule 
% Fast and Furious (FaF)~\cite{Luo18CVPR}. &  &  &  &  &  &  &  &  &  &  &  &  &  &  &  &  &  &  &  &  &  &  & \tabularnewline
% %
% Our+precast (post NMS) &  &  &  &  &  &  &  &  &  &  &  &  &  &  &  &  &  &  &  &  &  &  & \tabularnewline
% \midrule 
% Our+precast (pre NMS) &  &  &  &  &  &  &  &  &  &  &  &  &  &  &  &  &  &  &  &  &  &  & \tabularnewline
% \midrule 
% Our + bi-directional &  &  &  &  &  &  &  &  &  &  &  &  &  &  &  &  &  &  &  &  &  &  & \tabularnewline
% \midrule 
% Our+conditioned &  &  &  &  &  &  &  &  &  &  &  &  &  &  &  &  &  &  &  &  &  &  & \tabularnewline

\caption{Joint car detection and forecasting evaluation on nuScenes. We adopt top-K evaluation for forecasting and evaluate under two settings of $K=1$ and $K=5$ (for forecasting only). We further breakdown the performance of each model by examining the detection AP ($AP_{det.}$) and forecasting AP ($AP_f$) on static, linear, and non-linearly moving sub-categories. First, we find that methods that are trained to detect objects in the current frame have higher overall $AP_{det.}$ (Detection + Constant Velocity, row 1), while methods that are trained to detect objects in future frames have higher overall $AP_f$ (c.f. \futuredet, row 4), which is expected by design. For forecasting, surprisingly, Trajectron++ (row 3) is outperformed by constant velocity predictions (row 1), suggesting that this is indeed a challenging problem and constant velocity is a strong baseline. \futuredet consistently outperforms other baselines on non-linear trajectories. Notably, for $K=5$, we improve the non-linear object forecasting accuracy by 4\% over FaF*. \futuredet trained with a PointPillars backbone provides modest improvement across metrics, and performs best overall. } %\deva{We don't define base AP. Should we just remove the firsrt 2 cols?}} 
\label{table:results}
\end{table*}

% Constant position (CP).
% Constant velocity (CV).
% Fast and Furious (FaF)~\cite{Luo18CVPR}.
% } 
% Our+precast+post-NMS:} 
% Our+bi-directional:} 
% Our+conditioned:}

After confirming that our proposed \metric is a suitable metric for joint object detection and forecasting, we compare \textit{FutureDet} to a number of baselines and two state-of-the-art methods.

%\subsection{Baselines}

% \PARM{Detection + Constant Position (CP).} This is our adaptation of CenterPoint that outputs identity detections and scores them inversely proportionally to the estimated velocity (Sec.~\ref{sec:breakdown}). This baseline assumes a static world and can only correctly forecast trajectories of objects that do not move. 
%
\PARM{Detection + Constant Velocity.} We begin with an surprisingly simple, yet strong baseline which is often overlooked in forecasting literature. This baseline takes the detections, as well as the estimated velocity from our CenterPoint detector \cite{yin2021center}, and simply extrapolated forecasts as if objects are moving with constant velocity. Since CenterPoint is such a strong detector, this baseline produces strong results. Most of the ground-truth objects are approximately moving with a constant velocity, either moving directly forward or are stationary. We expect this model to under-perform on non-linear trajectories. 

% TDifferent to CP model, this model simply stacks the velocity prediction, assuming objects move linearly with a constant velocity. 
%
\PARM{Detection + Forecast (FaF$^{*}$, \cf~\cite{Luo18CVPR}).} This variant predicts a different velocity offset at every time step into the future for each detection, and derives trajectories by integrating velocities in the forward direction. This is precisely what Fast and Furious (FaF) does~\cite{Luo18CVPR}. For an apples-to-apples comparison, we re-implement FaF using a CenterPoint-backbone and denote this model as FaF$^{*}$. This method predicts a single trajectory per  detection. 
%
%(Sec.~\ref{sec:methodology}). Does not use current detections. Cannot do multi-future.
%

\begin{figure*}[ht!]
\begin{center}
    \includegraphics[width=0.99\linewidth,  trim= 0.5cm 2.5cm 1cm 0cm, clip]{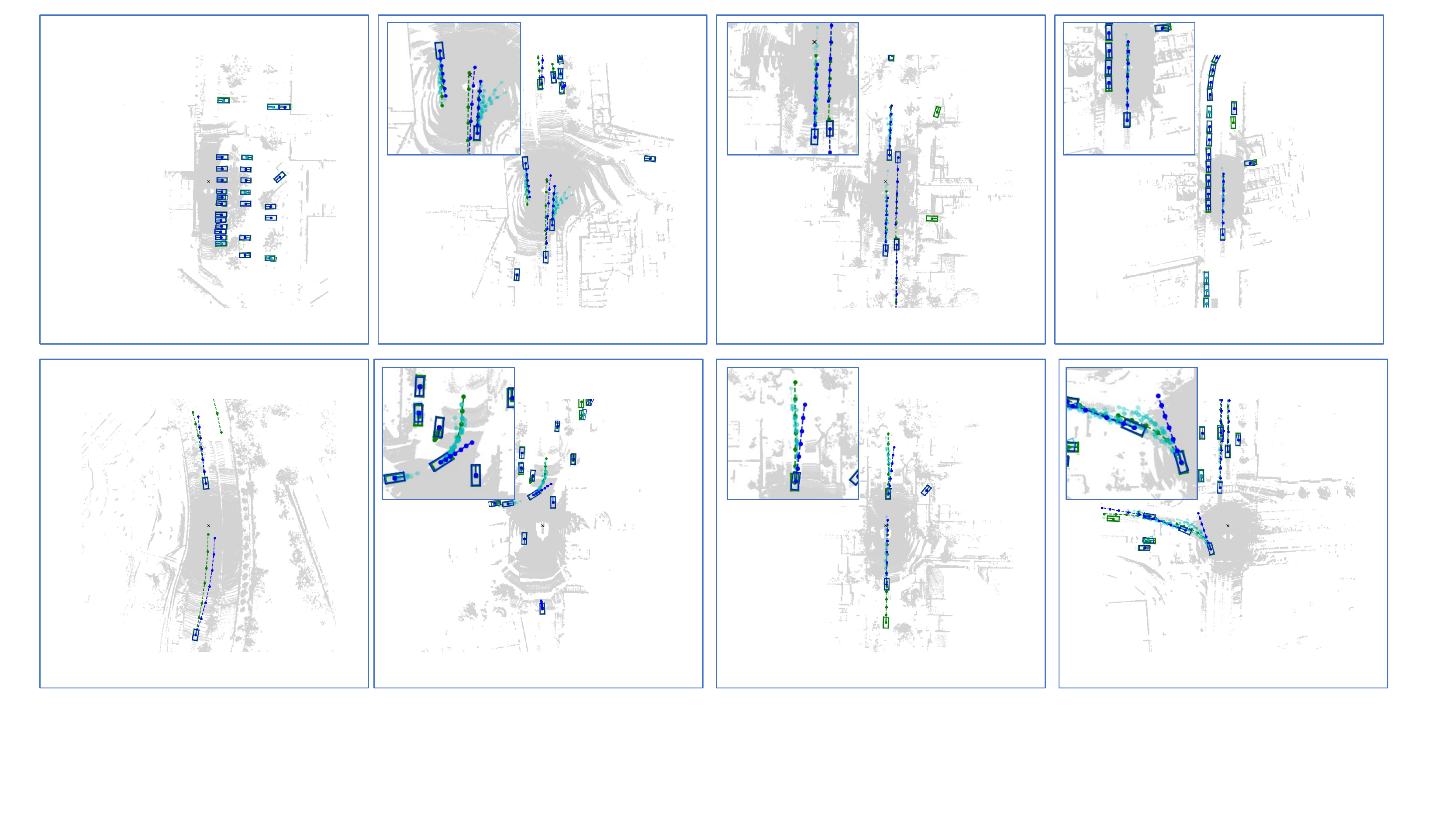}
\end{center}
\caption{We qualitatively evaluate forecasts from \futuredet. We denote ground-truth trajectories with \textbf{\textcolor{darkgreen}{green}} and multiple future predictions with \textbf{\textcolor{blue}{blue}} for the highest confidence forecast and \textbf{\textcolor{cyan}{cyan}} for the remaining multiple-future predictions.  Since we repurpose CenterPoint, a state-of-the-art detector, current frame detection performs well. Often, our model predicts that moving objects may be moving with constant velocity with high confidence. Given the data bias, where most objects are either stationary or are moving with constant velocity, this is a reasonable output. We highlight the multiple-future detection output in the top left.}
\label{fig:qualitative}
\end{figure*}

\PARM{Trajectron++.} We compare all of the aforementioned variants and ablations of our method to the state-of-the-art auto-regressive trajectory prediction model, Trajectron++~\cite{salzmann2020eccv}.  This model is indicative of current state-of-the-art approaches for the traditional forecasting task where ground truth tracks are given. With this comparison, we wish to outline how the standard three-stage detection-tracking-forecasting approach compares with our end-to-end forecasting method. To construct this baseline, we begin with off-the-shelf detection and tracking results from CenterPoint \cite{yin2021center}. CenterPoint performs tracking using the velocity offset estimates to match detections in each frame using a greedy matching between the current frame detections and previous frame detections. Trajectron++ then takes these predicted trajectories as input for forecasting. 
%For $K=1$ we generate the single most likely forecast.  For $K=5$, we also sample four more highly confident trajectories. % We give the mode forecast \alj{???} a slightly higher forecasting score than the other samples. This was the multi-future setting that worked best. When just randomly sampling $5$ forecasts without taking the mode forecast, the results ended up worse than just taking the mode forecast. \alj{Everything after ??? makes no sense to me (maybe its the 1.48am but check).}
\PARM{\futuredet.}. We detect objects directly in future frames and backcast these future detections to the reference frame. Intuitively, the advantage of this variant over simple forecasting (FaF) is in that it encourages the network to learn a better feature representation for forecasting by placing ''multiple bets'' on the future position of objects in the current frame. As shown in Figure \ref{fig:method}, this method naturally allows for a multiple-future interpretation of the observed sensory data (as discussed in Sec.~\ref{sec:methodology}). In Figure \ref{fig:qualitative}, we show qualitatively that our method can represent multiple futures. We note that the highest confidence future trajectories looks like constant velocity predictions as the training data is biased towards static and linearly moving objects. \futuredet is able to learn road geometries without map information, as indicated by the curved trajectories. 

%\PARM{\futuredet} This entry denotes our full proposed method: we detect objects in current and future frames, backcast and match possibly multiple future detections to the set of current frame detections. This variant naturally provides a multi-future interpretation of the observed sensory data (as discussed in Sec.~\ref{sec:methodology}). In Fig. \ref{fig:qualitative} we show qualitatively that our method can represent multiple futures. We note that the most confident future trajectory looks like a constant velocity prediction as the training data is biased towards static and linearly moving objects. \futuredet is able to learn road geometries without map information, as indicated by the curved trajectories. 

\PAR{Discussion.}
We compare the results of the aforementioned variants to \futuredet as well as Trajectron++~\cite{salzmann2020eccv} in Table~\ref{table:results}. First, we notice that moving object forecasting under our end-to-end setting is a challenging problem --- none of the methods we study have high $AP_f$, suggesting the need for the community to focus on this problem. Second, despite the constant velocity model being conceptually simple, it performs on par with our FaF re-implementation and improves on Trajectron++ by +7.8 $mAP_f$. Unfortunately, this constant velocity baseline is usually under-emphasized in the literature. We argue here that it still serves as an important baseline. The poor performance of Trajectron++ might also hint that performing direct end-to-end forecasting is advantageous over a three-stage approach of detection-tracking-forecasting, where errors can easily compound.
\futuredet takes a different approach compared to existing methods. Our method improves upon all other baselines in terms non-linear object $AP_f$ and the motion category-averaged $mAP_f$ (our primary metric). 
%Our first attempt of FutureDet + Backcast performs sub-par in terms of detection accuracy, especially in terms of moving object accuracy, suggesting that the matching step is critical. Matching removes a number of duplicated forecasts from being considered as completely different objects, and therefore restores the detection AP, and improves the forecasting $mAP$.
In addition, this multi-future interpretation also allows the performance to be  improved in the $K=5$ evaluation, where the forecast with minimum FDE from the top 5 ranked forecasts for each detection is evaluated. Note that for \futuredet $AP_f^{static}$,  $K=1$ results slightly decrease because bundling multiple trajectory estimates into one multi-future prediction for a single object reduces recall. However this is more than made up for in the increase in performance for detection and forecasting moving objects at $K=5$.
We train a version of our model with road masks as an additional input channel into the BEV feature representation (after the sparse-voxel backbone). This brings very little change to the results. We hypothesize that adding the map information does not provide additional information as it can be easily be learnt from the raw LiDAR input. However, further exploration is required to evaluate how to  best fuse map information.

\section{Conclusion}
This paper presents a new end-to-end method for trajectory forecasting directly from LiDAR sensor data. Our proposed \futuredet is a natural forecasting-by-detection framework that allows for a multi-future interpretation of the observed evidence and establishes a new state-of-the-art. 
We provide thorough analysis of existing evaluation metrics for end-to-end forecasting and reveal that they can be gamed by a simple constant position model. To this end, we propose a new set of evaluation metrics based on the average precision metric that comprehensively evaluates joint detection and forecasting performance. This allows us to conduct a thorough analysis that reveals that a constant velocity model is a surprisingly strong baseline that should be considered in future forecasting work.  

%framework that can predict multiple future trajectories. We reframe forecasting as the task of future detection, naturally facilitating multiple future predictions.

\PAR{Limitations.}
%Prior work in classic trajectory prediction methods not only predict multiple-future trajectory estimates but also generate diverse multi-modal predictions. 
As we do not explicitly enforce diverse trajectory generation, many of our multiple-futures are closely clustered. While \futuredet presents the first method for end-to-end forecasting from raw sensory data, capable of multi-future predictions, generation of diverse, multi-modal predictions remains an open challenge. 

% Working directly with sensor data means that the distribution of potential trajectories is often uninteresting. Most objects in a scene are either stationary, or moving with constant velocity. We solicit contributions for future datasets that explicitly address this challenge. 

\PAR{Acknowledgments.}
This work was supported by the CMU Argo AI Center for Autonomous Vehicle Research.

\clearpage
%%%%%%%%% REFERENCES
{\small
\bibliographystyle{ieee_fullname}
\bibliography{abbrev_long,refs}
}
\clearpage 

\appendix

\section{Examining \futuredet's Predictions}
\begin{figure}[bt!]
\begin{center}
    \includegraphics[width=1.0\linewidth,  trim= 0cm 3cm 18cm 0cm, clip]{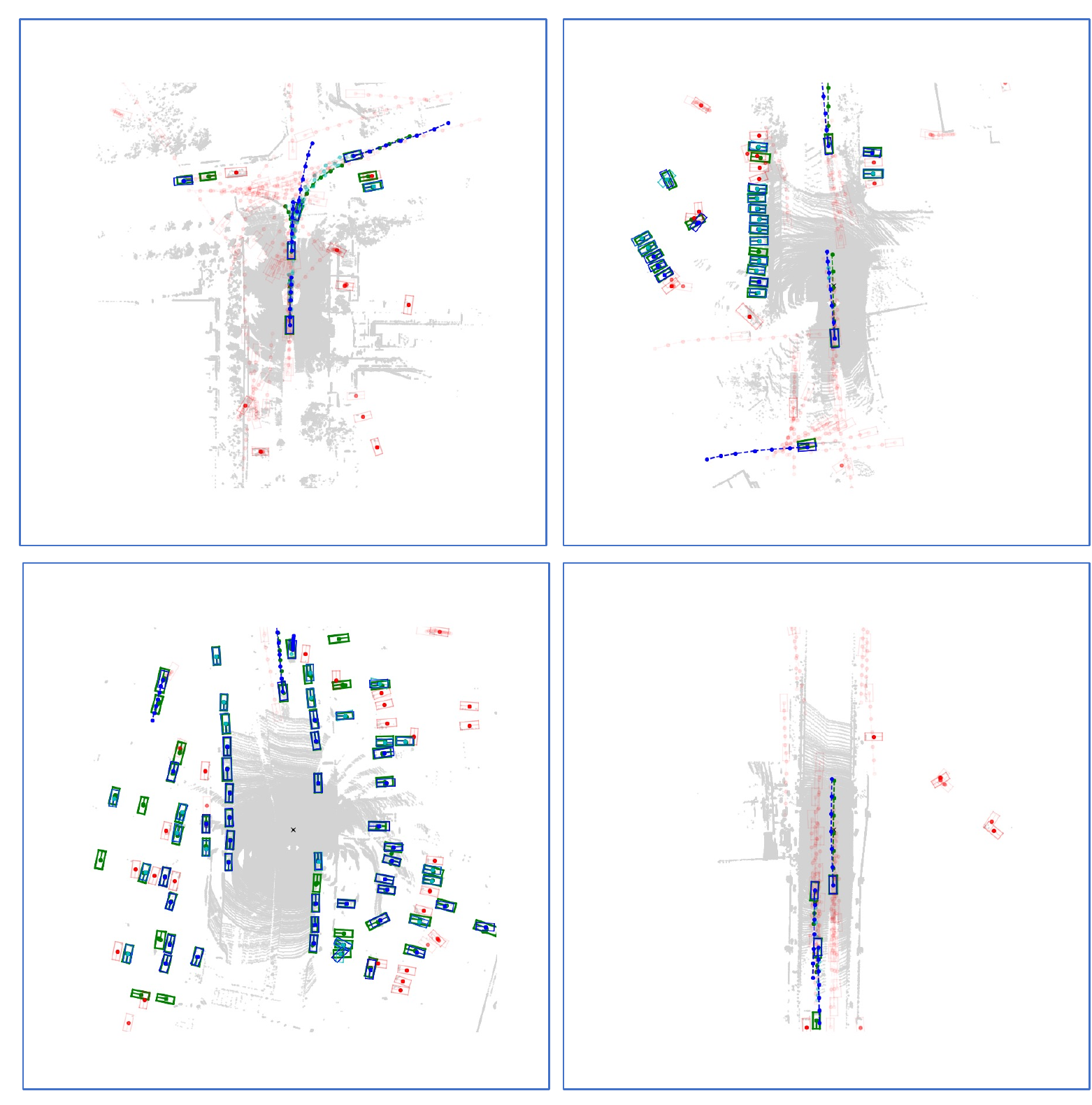}
\end{center}
\caption{The raw output of \futuredet includes many false positive detections and forecasts (shown in \textbf{\textcolor{red}{red}}). Further post-processing is required to leverage the output of our end-to-end model in further downstream tasks.}
\label{fig:qualitative_fp}
\end{figure}

One of the challenges of forecasting from raw sensor data is appropriately handling false positive detections and forecasts. The standard forecasting setup allows us to build models in isolation from other factors. However, the standard assumption of having perfect input trajectories is not feasible in practice as it critically depends on \textit{perfect} object tracks (and by extension perfect detections) as inputs, which are nearly impossible to obtain in practice. As seen in Figure \ref{fig:qualitative_fp}, the raw output of \futuredet makes it challenging to use in practice. Intuitively, training a model to make ``multiple bets'' about the position of objects may induce more false positives. Further post-processing is required to leverage the output of our end-to-end model in further downstream tasks.

\section{Evaluating Pedestrian Forecasting}
In this section, we evaluate pedestrian forecasting on the nuScenes dataset. Forecasting pedestrian movement can be considerably more challenging than car forecasting because pedestrians are more dynamic. Given our 3 second forecasting horizon, pedestrians typically do not move very far from their initial position. As a result, we define tighter match thresholds for pedestrian forecasting.
A successful match in the current frame is determined based on the distance from the center, averaged over distance thresholds of $\{0.125, 0.25, 0.5, 1\} m$. A successful match in the final timestep is determined based on the distance from center, averaged over distance thresholds of $\{0.25, 0.5, 1, 2\} m$ respectively. We highlight the results of pedestrian forecasting in Table \ref{tab:results_ped}.

\begin{figure}[ht!]
\begin{center}
    \includegraphics[width=1.0\linewidth,  trim= 0.0cm 3cm 17.7cm 0cm, clip]{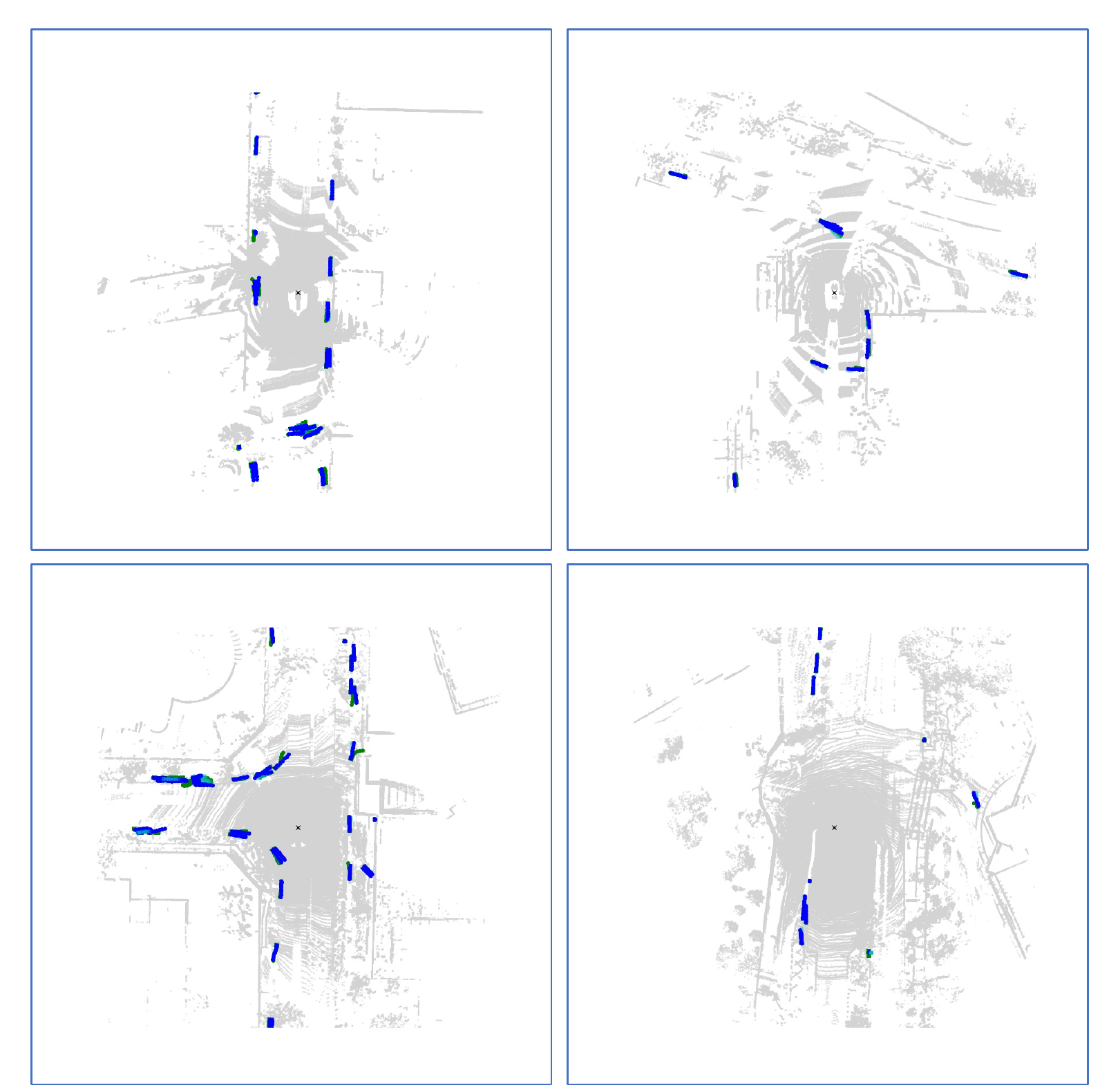}
\end{center}
\caption{We qualitatively evaluate pedestrian forecasts from \futuredet (we denote the ground-truth trajectories with \textbf{\textcolor{darkgreen}{green}} and multiple future predictions with \textbf{\textcolor{blue}{blue}} for the highest confidence forecast and \textbf{\textcolor{cyan}{cyan}} for the remaining future predictions). Pedestrian forecasting is more difficult than car forecasting due to the dynamic movement of pedestrians. \futuredet struggles to accurately forecast pedestrians because they often travel in crowds. This makes it difficult to accurately detect and forecast individual pedestrian motion. Often, the predicted multiple futures are all linearly moving, and are often similar to each other. }
\label{fig:qualitative_ped}
\end{figure}

We see that \futuredet performs the best overall, with 26.9 $mAP_f$. Looking to Figure \ref{fig:qualitative_ped}, it is clear that pedestrian detections are tightly clustered together, making back-casting less effective overall. We also find that many of the predicted multiple-futures are very similar to one another, indicating that the model is not able to model dynamic pedestrian futures. However, \futuredet still consistently improves over FaF* by 1\% on $AP_f$ metrics. %As with car forecasting, we find that Trajectron++ performs worse (\neehar{XXX} $mAP_f$) than all other models.

We train a version of our model with road masks as an additional input channel into the BEV feature representation (after the sparse-voxel backbone). This brings very little change to the results. We hypothesize that adding the map information does not provide additional information. However, further exploration is required to evaluate how to  best fuse LiDAR and map information.

\begin{table*}[ht!]
\centering
   % \scriptsize
%   \setlength{\tabcolsep}{0.8pt}
   \setlength{\tabcolsep}{01.5pt}
      \resizebox{\linewidth}{!}{%

\begin{tabular}{lccccccccccccccccccccccc}
\toprule 
 & \multicolumn{11}{c}{K=1} &  & \multicolumn{11}{c}{K=5}\tabularnewline
\cmidrule{2-12} \cmidrule{3-12} \cmidrule{4-12} \cmidrule{5-12} \cmidrule{6-12} \cmidrule{7-12} \cmidrule{8-12} \cmidrule{9-12} \cmidrule{10-12} \cmidrule{11-12} \cmidrule{12-12} \cmidrule{14-24} \cmidrule{15-24} \cmidrule{16-24} \cmidrule{17-24} \cmidrule{18-24} \cmidrule{19-24} \cmidrule{20-24} \cmidrule{21-24} \cmidrule{22-24} \cmidrule{23-24} \cmidrule{24-24} 
 & \multicolumn{2}{c}{$AP^{stat.}$} &  & \multicolumn{2}{c}{$AP^{lin.}$} &  & \multicolumn{2}{c}{$AP^{non-lin.}$} &  & \multicolumn{2}{c}{$mAP$} &  & \multicolumn{2}{c}{$AP^{stat.}$} &  & \multicolumn{2}{c}{$AP^{lin..}$} &  & \multicolumn{2}{c}{$AP^{non-lin.}$} &  & \multicolumn{2}{c}{$mAP$}\tabularnewline
\cmidrule{2-3} \cmidrule{3-3} \cmidrule{5-6} \cmidrule{6-6} \cmidrule{8-9} \cmidrule{9-9} \cmidrule{11-12} \cmidrule{12-12} \cmidrule{14-15} \cmidrule{15-15} \cmidrule{17-18} \cmidrule{18-18} \cmidrule{20-21} \cmidrule{21-21} \cmidrule{23-24} \cmidrule{24-24} 
 & $\textrm{AP}_{\textrm{det.}}$ & $\textrm{AP}_{\textrm{f}}$ &  & $\textrm{AP}_{\textrm{det.}}$ & $\textrm{AP}_{\textrm{f}}$ &  & $\textrm{AP}_{\textrm{det.}}$ & $\textrm{AP}_{\textrm{f}}$ &  & $\textrm{mAP}_{\textrm{det.}}$ & $\textrm{mAP}_{\textrm{f}}$ &  & $\textrm{AP}_{\textrm{det.}}$ & $\textrm{AP}_{\textrm{f}}$ &  & $\textrm{AP}_{\textrm{det.}}$ & $\textrm{AP}_{\textrm{f}}$ &  & $\textrm{AP}_{\textrm{det.}}$ & $\textrm{AP}_{\textrm{f}}$ &  & $\textrm{mAP}_{\textrm{det.}}$ & $\textrm{mAP}_{\textrm{f}}$\tabularnewline
\midrule 
% Detection + Constant Position (CP) & 62.0 & 54.0 & &  88.6 & 84.8  & & 3.6 & 0.0  & & 46.1 & 42.2  & & 62.0 & 54.0 & & 88.6 & 84.8 & & 3.6 & 0.0 & & 46.1 & 42.4  \tabularnewline
%\midrule 
Detection + Constant Velocity & \textbf{55.1} & 33.3 &  & 73.5 & 27.8 &  & 96.9 & 12.4 &  & \textbf{75.2} & 24.5  &  & \textbf{55.1} & 33.3 &  & 73.5 & 27.8 &  & 96.9 & 12.4 &  & \textbf{75.2} & 24.5 \tabularnewline
%\midrule 
Detection + Forecast (\cf~\cite{Luo18CVPR}) & 53.7 & \textbf{35.0} &  & \textbf{73.9} & 30.8 &  & \textbf{97.2} & 13.3 &  & 74.9 & 26.4  &  & 53.7 & 35.0 &  & \textbf{73.9} & 30.8 &  & \textbf{97.2} & 13.3 &  & 74.9 & 26.4 \tabularnewline

Trajectron++~\cite{salzmann2020eccv}  & \textbf{55.1} & 16.4 &  & 73.5 & 7.8 &  & 96.9 & 5.2 &  & \textbf{75.2} & 9.8  &  & 55.1 & 18.1 &  & 73.5 & 9.0 &  & 96.9 & 6.9  &  & \textbf{75.2} & 11.3 \tabularnewline
\midrule 

\rowcolor{green!10} \textbf{\futuredet}  & 53.1 & 33.3 &  & 72.4 & \textbf{32.6 }&  & 95.3 & \textbf{14.7}  &  & 73.6 & \textbf{26.9}  &  & 53.1 & \textbf{35.1} &  & 72.4 & \textbf{34.0} &  & 95.2 & \textbf{15.0}  &  & 73.6 & \textbf{28.0} \tabularnewline
\rowcolor{green!10} \futuredet-PointPillars  & 41.0 & 20.7 &  & 69.1 & 29.8 &  & 93.3 & 13.3  &  & 67.8 & 21.3  &  & 41.0 & 22.9 &  & 69.2 & 31.0 &  & 93.1 & 13.5  &  & 67.7 & 22.5 \tabularnewline
\rowcolor{green!10} \futuredet + Map  & 52.4 & 33.0 &  & 71.8 & 32.0 &  & 95.3 & 14.4  &  & 73.2 & 26.5  &  & 52.4 & 34.8 &  & 71.8 & 33.4 &  & 95.2 & 14.8  &  & 73.2 & 27.7 \tabularnewline

% \midrule 
\bottomrule
\end{tabular}
}%

%
% Constant position (CP) &  &  &  &  &  &  &  &  &  &  &  &  &  &  &  &  &  &  &  &  &  &  & \tabularnewline
% \midrule 
% Constant velocity (CV) &  &  &  &  &  &  &  &  &  &  &  &  &  &  &  &  &  &  &  &  &  &  & \tabularnewline
% \midrule 
% Fast and Furious (FaF)~\cite{Luo18CVPR}. &  &  &  &  &  &  &  &  &  &  &  &  &  &  &  &  &  &  &  &  &  &  & \tabularnewline
% %
% Our+precast (post NMS) &  &  &  &  &  &  &  &  &  &  &  &  &  &  &  &  &  &  &  &  &  &  & \tabularnewline
% \midrule 
% Our+precast (pre NMS) &  &  &  &  &  &  &  &  &  &  &  &  &  &  &  &  &  &  &  &  &  &  & \tabularnewline
% \midrule 
% Our + bi-directional &  &  &  &  &  &  &  &  &  &  &  &  &  &  &  &  &  &  &  &  &  &  & \tabularnewline
% \midrule 
% Our+conditioned &  &  &  &  &  &  &  &  &  &  &  &  &  &  &  &  &  &  &  &  &  &  & \tabularnewline

\caption{Joint pedestrian detection and forecasting evaluation on nuScenes. We adopt top-K evaluation for forecasting and evaluate under two settings of $K=1$ and $K=5$ (for forecasting only). We further breakdown the performance of each model by examining the detection AP ($AP_{det.}$) and forecasting AP ($AP_f$) on static, linear, and non-linearly moving sub-categories. Note that since pedestrians have smaller displacement over a 3 second forecasting horizon, we tighten the match thresholds as described above. \futuredet performs the best, improving over FaF* by 0.5 $mAP_f$. As with car forecasting, FaF* and the constant velocity baseline beat Trajectron++ by 14.4 \% and 16.6 \% $mAP_f$ respectively. Notably, training with a PointPillars backbone dramatically reduces \futuredet performance on all metrics. In addition, we find that using a road mask does not significantly change the performance of \futuredet, indicating that the model might already be reasoning about spatial context.} %\deva{We don't define base AP. Should we just remove the firsrt 2 cols?}} 
\label{tab:results_ped}
\end{table*}

% Constant position (CP).
% Constant velocity (CV).
% Fast and Furious (FaF)~\cite{Luo18CVPR}.
% } 
% Our+precast+post-NMS:} 
% Our+bi-directional:} 
% Our+conditioned:}

\section{\futuredet  Architecture}
In this section, we further describe the implementation details of \futuredet. Specifically, we focus on the detector head architecture, and the sampling strategy used to improve nonlinear trajectory forecasting. 

\begin{figure}[ht!]
\begin{center}
    \includegraphics[width=1.0\linewidth,  trim= 0cm 5cm 0cm 2cm, clip]{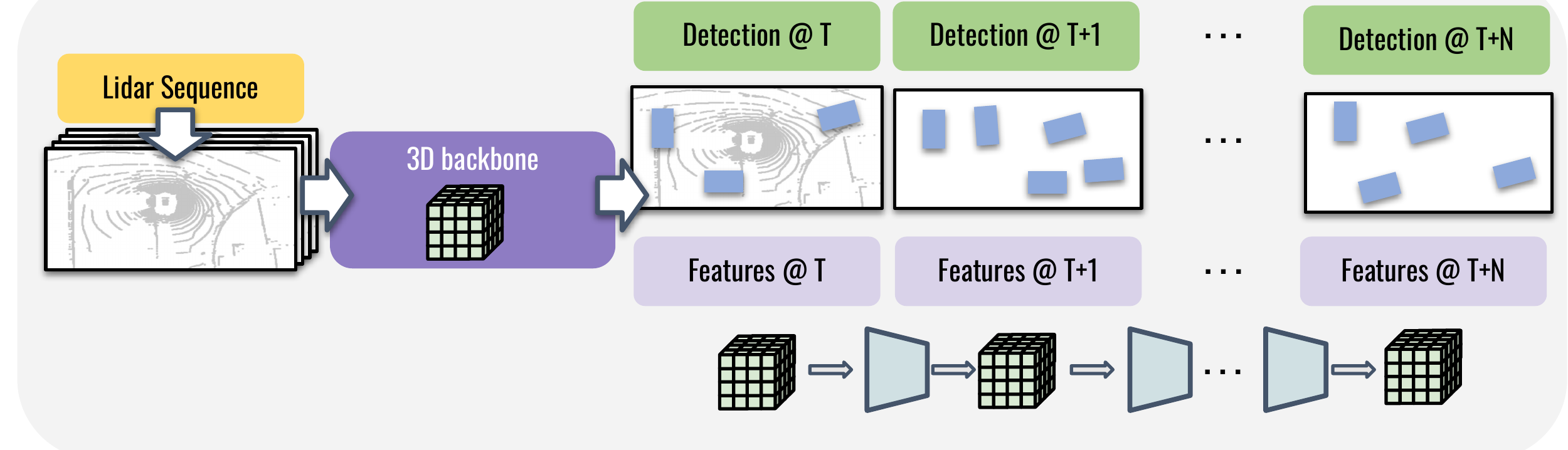}
\end{center}
\caption{\futuredet's detector head architecture adapts CenterPoint's architecture for the task of forecasting. Importantly, CenterPoint shares one set of features for all classes (i.e. cars, trucks, pedestrians, etc.). Since we adapt the architecture to forecast {\tt cars} and {\tt future-cars}, the single shared feature may not be able to effectively model long-term forecasting. To this end, we allow the model to learn a shallow network that transforms current features into future features.}
\label{fig:arch}
\end{figure}

\textbf{Recurrent Features}. We re-purpose CenterPoint for our implementation of \futuredet. However, CenterPoint is designed to detect objects in the current frame. It uses a shared feature representation for all classes. Although this effectively captures object spatial location, it does not allow for a robust representation of forecasted features. Specifically, since \futuredet detects {\tt cars} and {\tt future-cars}, we expect that the features required to detect these temporally offset classes should be different. To this end, we allow the model to learn a shallow network that transforms current features into future features as shown in Figure \ref{fig:arch}. 
%Our ablation study (Table \ref{tab:ablation_recurrent}) suggests that these recurrent features slightly improve \futuredet's ability to forecast at $K=1$ and significantly improves at $K=5$.

\textbf{Trajectory Sampler}. The distribution of static, linear, and non-linear trajectories in the nuScenes dataset is unbalanced. Since most cars are parked, we find that 60\% of the trajectories are static. In order to address this data imbalance, we leverage copy-paste augmentation proposed by \cite{zhu2019class} to oversample linear and nonlinear trajectories during training. Importantly, we ensure that our copy-paste augmentation samples at the trajectory level, instead of at the class level, allowing consistent augmented trajectories across all detection heads (i.e. classes). 

\section{Computing Motion Subclass AP}
Computing subclass average precision is straightforward in principle if both predictions and ground-truth have subclass labels; one can simply treat the sub-class as a class and apply standard precision-recall metrics. In our case, predictions do not come with a subclass label. Instead, we match predictions to ground-truth at a class-level, and assign the ground-truth sub-class to the true positive matched predictions. However, this will not produce any sub-class labels for false positive predictions (that are unmatched). Instead, the metric evaluation code {\em derives} sub-class labels for false positive predictions, by applying the same logic used to derive sub-class labels for the ground-truth. We follow this procedure as it is also used to produce small-vs-large sub-class precision-recall metrics for standard detection toolkits~\cite{lin2014microsoft}. Finally, although we use the language of sub-classes, our formalism can apply to any attributes associated with a detection. 
%\deva{Could replace subclass with attribute; need to be consistent in draft and appendix.}

We derive the subclass label as a function of the (ground truth or predicted) trajectory. For each trajectory, we first compute the IoU between bounding boxes at the first and last timestep. If the IoU is greater than 0, this trajectory is considered to be static. Next, using the velocity of the first timestep bounding box, we apply a constant velocity forecast to the initial position to compute a target box. If the IoU between the last timestep box and target box is greater than 0, this trajectory is considered to be linear. All trajectories that are not classified as static or linear as considered to be non-linear trajectories. 

\section{Broader Impact}
Autonomous agents will play an important role in the automation of tasks that can be considered unsafe (\eg, due to a high number of traffic accidents). Forecasting is at the heart of autonomous vehicle navigation: safe navigation necessitates motion prediction of surrounding agents to ensure driving safety. By leveraging LiDAR sensory data to accomplish this task, we can better understand world geometry and dynamics. Moreover, establishing the proper metrics, particularly considering the performance of moving and static car trajectories, is essential for building safe embodied robotics systems.

\end{document}